\pdfoutput=1

\documentclass[11pt]{article}

\usepackage[]{naacl2021}

\usepackage{times}
\usepackage{latexsym}

\usepackage[T1]{fontenc}

\usepackage[utf8]{inputenc}

\usepackage{microtype}

\usepackage{adjustbox}
\usepackage{booktabs}
\usepackage{multirow}
\title{IndT5: A Text-to-Text Transformer for 10 Indigenous Languages}

\vspace{3mm}
\author{El Moatez Billah Nagoudi$^1$, Wei-Rui Chen$^1$,  Muhammad Abdul-Mageed$^1$,  Hasan Cavusoglu$^2$ \\
$^1$ Natural Language Processing Lab,\\ 
      $^{1,2}$ The University of British Columbia  \\
     \small{$^1$ \{moatez.nagoudi,weirui.chen,muhammad.mageed\}@ubc.ca, $^2$ cavusoglu@sauder.ubc.ca}}

\begin{document}
\maketitle
\begin{abstract}
Transformer language models have become fundamental components of natural language processing based pipelines.  Although several Transformer models have been introduced to serve many languages, there is a shortage of models pre-trained for low-resource and Indigenous languages.  In this work, we  introduce IndT5, the first Transformer language model for Indigenous languages. To train IndT5, we build IndCorpus--a new dataset for ten Indigenous languages and Spanish. We also present the application of  IndT5 to machine translation by investigating  different approaches to translate between Spanish and the Indigenous languages as part of our contribution to the AmericasNLP 2021 Shared Task on Open Machine Translation.  IndT5 and IndCorpus are publicly available for research.\footnote{\url{https://github.com/UBC-NLP/IndT5}}\vspace{2mm}
\end{abstract}

\section{Introduction}\label{sec:introduction}
\vspace{2mm}
Indigenous languages are starting to attract attention in the field of natural language processing (NLP), with the number of related publications growing in recent years \cite{mager-etal-2018-challenges}. In spite of this interest, there remains a multitude of challenges for handling Indigenous languages. Complexity of the morphological systems of some of these languages and lack of standard orthography for writing them are among these challenges~\cite{mager-etal-2018-challenges,  littell-etal-2018-indigenous}. The most fundamental issue facing NLP efforts, however, remains the lack of digital textual data that can be exploited for systems development.


In this work, we describe a scenario usually faced when trying to develop NLP systems for Indigenous languages and we focus on machine translation (MT). We adopt a neural machine translation approach (NMT)~\cite{koehn2017neural} as our method. We show that, in spite of its recent success on many contexts, NMT still struggles in very low-resource settings involving Indigenous languages. This is due to the core difficulty of lack of parallel textual data, but also even monolingual data. 


\begin{figure}[t]
\begin{centering}
\includegraphics[scale=0.30]{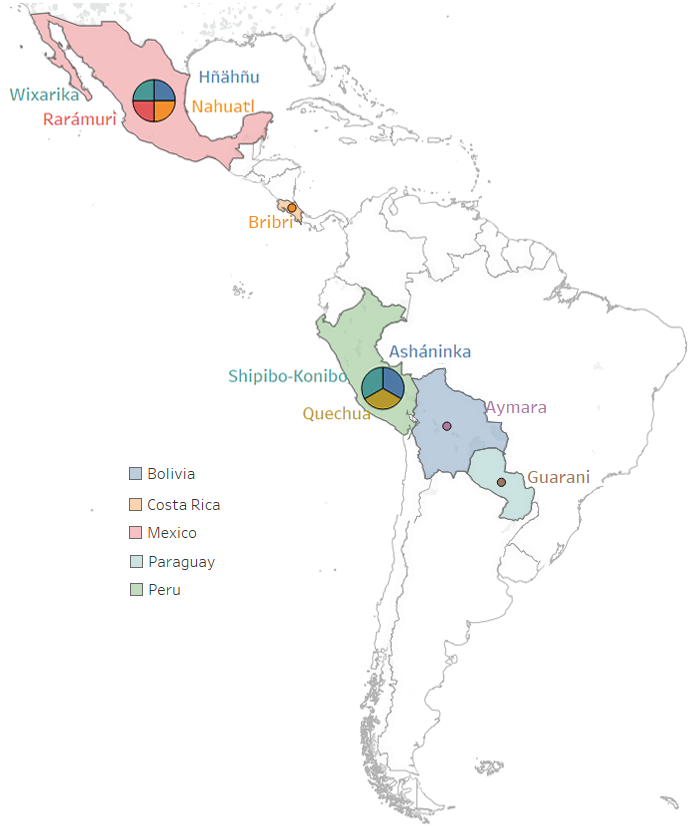}
  \caption{A map of the ten Indigenous languages covered by IndT5, our text-to-text Transformer model, and our IndCorpus dataset. The languages are mainly spoken in five Latin American countries.}
  \label{fig:cities}
 \end{centering}
\end{figure}

Although our main goal in this work in particular is to develop translation models from Spanish to several Indigenous languages of the Americas, we adopt a transfer learning approach where we offer resources that can be exploited for other downstream tasks. Namely, we build a  dataset for ten Indigenous languages and Spanish which we refer to as \textbf{\texttt{IndCorpus}}.  Figure~\ref{fig:cities} and Table~\ref{tab:langs_overview} provide an overview of the ten Indigenous languages in our new dataset~\cite{Ethnologue2021}.  We also exploit \textbf{\texttt{IndCorpus}} for pre-training a Transformer language model following the unified approach introduced by~\cite{raffel2019exploring}. Our resulting model, \textbf{\texttt{IndT5}}, treats every text NLP problem as a ``text-to-text" problem, i.e. taking text as input and producing new text as output. We apply \textbf{\texttt{IndT5}} to the MT task as a way to transfer knowledge acquired by the model to this particular context. Our experiments show the utility of our new language model and the dataset it exploits for the downstream Indigenous MT task but that very large space for improvement still exists.


\begin{table}[t]
\centering
\resizebox{1\columnwidth}{!}{%
 \begin{tabular}{llcr}
 \toprule
\textbf{Language} & \textbf{Code}  & \textbf{Main location}  & \textbf{Speakers} \\
\toprule
Aymara          & aym  & Bolivia   & 1,677,100 \\ 
Asháninka       & cni  & Peru  & 35,200          \\ 
Bribri          & bzd  & Costa Rica   & 7,000   \\
Guarani         & gn   & Paraguay & 6,652,790   \\ 
Hñähñu          & oto  & Mexico   & 88,500 \\
Nahuatl         & nah  & Mexico   & 410,000   \\
Quechua         & quy  & Peru   & 7,384,920   \\
Rarámuri        & tar  & Mexico   & 9,230   \\
Shipibo-Konibo  & shp  & Peru   &  22,500   \\
Wixarika        & hch  & Mexico   & 52,500  \\
\toprule 
\end{tabular}}
\caption{Overview of our ten Indigenous languages ~\cite{Ethnologue2021}.}  \label{tab:langs_overview}

\end{table}
\normalsize



The rest of the paper is organized as follows: In Section~\ref{sec:literature}, we introduce recent MT work in low-resource and Indigenous languages settings. In Section~\ref{sec:LM}, we describe how we develop our new language model for ten Indigenous languages. In Section~\ref{sec:our_MT_models}, we describe our NMT models. We conclude in Section~\ref{sec:conc}.

\section{Related Work}\label{sec:literature}


\subsection{Low-Resource MT}
A number of methods and techniques have been proposed to mitigate the effects of having rather small datasets for machine translation. These include data augmentation, transfer learning, hyperparameter tuning, incorporating linguistic knowledge, and knowledge distillation.

Since the main bottleneck of low-resource MT is the lack of abundant parallel textual data, data augmentation is straightforwardly a potential method to enhance the model performance. Back translation is a way to augment parallel data \cite{sennrich-etal-2016-improving}. By training a target-to-source translation model with original data and feeding in monolingual data of target language, synthetic parallel data is generated. If the target language is rich in textual data, much synthetic parallel data can be added into training data and may benefit the final translation model.

Transfer learning is another method that can boost the performance of MT on low-resource languages \cite{zoph2016transfer,DBLP:journals/corr/abs-1708-09803,DBLP:journals/corr/abs-1809-00357}. The rationale behind one approach to transfer learning is that knowledge obtained while translating high-resource languages may be transferable to translation of low-resource languages. In \newcite{zoph2016transfer}, a parent model is first trained on a high-resource language pair (i.e., French to English) then a child model is trained on a low-resource language pair (i.e., Uzbek to English). The Uzbek-English model has $10.7$ BLEU score without parent model and $15.0$ with the parent model. It is also shown that the more similar the two source languages, the more performance gain is possible. For example, a Spanish-English MT model has $16.4$ BLEU score without parent model and $31.0$ with French-English parent model. The performance gain is much more than when transferring French-English parent model to the more distant context of the Uzbek-English child model.


\newcite{sennrich2019revisiting} argue that instead of using hyperparameters that work in high-resource settings, there should be a set of hyperparameters specific to the low-resource scenario. For example, keeping the vocabulary size small, training a model with relatively small capacity, and having smaller batch size may be beneficial to model performance. When building a vocabulary with BPE, by reducing the the number of merge operations, a smaller vocabulary can be obtained and an inclusion of low-frequency (sub)words can be avoided. Inclusion of inclusion of low-frequency (sub)words could otherwise negatively influencing representation learning effectiveness.


Leveraging linguistic knowledge for data augmentation,~\newcite{zhou2019handling} use a rule-based syntax parser and a dictionary to generate parallel data. By reordering target-language sentences into source-language syntactic structure and then mapping target-language words into source-language words with a dictionary, the size of  parallel data is enlarged and translation performance is improved.

\newcite{baziotis2020language} leverage a language model to help enhance the performance of the translation model. Similar to the idea of knowledge distillation \cite{hinton2015distilling}, a teacher model and a student model are trained where the language model plays the role of teacher and translation model plays the role of student. With this design, the teacher model needs only monolingual data and does not have to rely on large parallel data.

\subsection{MT of Indigenous Languages}

Unlike high-resource languages such as English and French, Indigenous languages are often low-resource. Due to this, it is common that researchers of Indigenous languages adopt methods that can fare well in low-resource scenarios. This includes using the Transformer architecture and its variants in both low-resource~\cite{adebara2021translating,adebara-etal-2020-translating,przystupa2019neural} and Indigenous language~\cite{feldman2020neural,orife2020neural, Tan2020AddressingCO} settings. 

Despite the fact that Indigenous languages face difficulties similar to most low-resource languages, there are some challenges specific to Indigenous languages. As~\newcite{mager-etal-2018-challenges} point out, some Indigenous languages have complex morphological systems and some have various non-standardized orthographic conventions. For example, \newcite{micher2018addressing} shows that in Inuktitut, an Indigineous language in North America with a complex morphological system, a corpus of one million tokens, there are about 225K different types for Inuktitut while about 30K types for English. Also, \newcite{micher2018addressing} shows that there can be lack of standardized spelling for some words. For example, the word \textit{Haammalat} in Inuktitut has another seven different forms.

To cope with the issue of complex morphology,~\newcite{Ortega-et-al-2020} build a translation model for Qeuchua, an Indigenous language of South America, with an integrated morphological segmentation method. To treat orthographic variation, \newcite{feldman2020neural} standardize text with a rule-based system which converts diacritics and letters to contemporary orthographic convention.

\begin{table*}[ht]
\centering

\begin{tabular}{lcccc}
\toprule
\multicolumn{1}{c}{\multirow{2}{*}{\textbf{Target language}}} & \multicolumn{2}{c}{\textbf{Wikipedia }} & \multicolumn{2}{c}{\textbf{ Bible}} \\  \cline{2-5}
  &   \textbf{ Size (MB)} & \textbf{Sentences} &  \textbf{Size (MB)} & \textbf{Sentences}   \\ \toprule

Hñähñu                   & -                &    -                             & 1.4     &    7.5K                                          \\
Wixarika                 & -            &       -                             &  1.3   &   7.5K
\\
Nahuatl                  & 5.8           &    61.1K                         & 1.5  &      7.5K

\\
Guarani                  & 3.7            &      28.2K                           & 1.3 &      7.5K                                              \\
Bribri                   & -               &    -                             & 1.5  &        7.5K                                        \\
Rarámuri                 & -                &     -                            & 1.9  &         7.5K                                       \\
Quechua                  & 5.9               &     97.3K                        & 4.9   &    31.1K                                            \\
Aymara                   & 1.7                &     32.9K                         & 5   & 30.7K
 \\
Shipibo-Konibo           & -                   &     -                         & 1    &    7.9K                                             \\
Asháninka                & -                    &     -                       & 1.4    &   7.8K                                          \\
Spanish                      & 1.13K             &    5M    & -              & - 
\\ \toprule
 
 \textbf{Total}  &  1.15K  &  5.22M   &    19.8 &     125.3K\\
 \toprule
\end{tabular}
\caption{Datasets in IndCorpus by language}\label{data_tab}
\end{table*}

\normalsize
\section{IndT5}\label{sec:LM}
\vspace{2mm}
We train an Indigenous language model adopting the unified and flexible text-to-text transfer Transformer (T5) approach~\cite{raffel2019exploring}. T5 treats every text-based language task as a ``text-to-text" problem, taking text format as input and producing new text format as output. T5 is essentially an encoder-decoder Transformer~\cite{vaswani2017attention}, with the encoder and decoder similar in configuration and size to a BERT\textsubscript{Base}\cite{devlin2019bert} but with some architectural modifications. Modifications include applying a normalization layer before a sub-block and adding a pre-norm (i.e., initial input to the sub-block output). We call our resulting model \textbf{\texttt{IndT5}}. We now describe our dataset, vocabulary, and pre-training method for developing \textbf{\texttt{IndT5}}.

\subsection{Training Data}
We build \textbf{\texttt{IndCorpus}}, a collection of ten Indigenous languages and Spanish comprising $1.17$ GB of text ($\sim$5.37M sentences), to pre-train \textbf{\texttt{IndT5}}. \textbf{\texttt{IndCorpus}} is collected from both Wikipedia and the Bible. Table~\ref{data_tab} provides the size and number of sentences for each language in our dataset.

\subsection{IndT5 Vocabulary}

The T5 \cite{raffel2019exploring} model is based on a vocabulary acquired by the SentencePiece library\footnote{\url{https://github.com/google/sentencepiece}} using English, French, German, and Romanian web pages from ``Colossal Clean Crawled Corpus" (or C4 for short). We use a similar procedure to create our Indigenous languages vocabulary. Namely, we use SentencePiece \cite{kudo2018subword} to encode text as WordPiece~\cite{smp2016} tokens with a vocabulary size of $100$K WordPieces extracted from \textbf{\texttt{IndCorpus}}.

\footnotesize 
\begin{table}[t]
\footnotesize
\centering
\resizebox{.9\columnwidth}{!}{%
 \begin{tabular}{lrcc}
 \toprule
\textbf{Languages} & \textbf{Train}  & \textbf{Dev}  & \textbf{Test} \\
\toprule
es-aym    & $6,531$ & $996$   & $1,003$   \\
es-cni  & $3,883$  & $883$   & $1,003$ \\
es-bzd   & $7,506$  & $996$   & $1,003$   \\
es-gn   & $26,032$ & $995$   &  $1,003$   \\
es-oto   & $4,889$ & $599$   & $1,003$ \\
es-nah    & $16,145$ & $672$   & $1,003$   \\
es-quy  & $125,008$  & $996$   & $1,003$   \\
es-tar   & $14,720$  & $995$   & $1,003$   \\
es-shp  & $14,592$ & $996$   &  $1,003$   \\
es-hch  & $8,966$ & $994$   & $1,003$  \\
\toprule 
\end{tabular}}
\caption{Distribution of MT data}  \label{tab:ST_data}

\end{table}
\normalsize

\subsection{Unsupervised Pre-Training}
We leverage our unlabeled Indigenous corpus, \textbf{\texttt{IndCorpus}}, to pre-train \textbf{\texttt{IndT5}}. For that, we use a denoising objective \cite{raffel2019exploring} that does not require labels. The main idea is  feeding the model with corrupted (masked) versions of the original sentence, and training it to reconstruct the original sentence.  Inspired by BERT's objective (i.e., masked language model)~\cite{devlin2019bert}, the  denoising objective ~\cite{raffel2019exploring} works by randomly sampling and dropping out $15$\% of tokens in the input sequence. All consecutive spans of dropped-out tokens are then replaced by a single sentinel token. We pre-train our model for $100$K steps on the \textbf{\texttt{IndCorpus}} using the T5\textsubscript{\textsubscript{Base}} architecture.\footnote{Both encoder and decoder of T5\textsubscript{Base} model has $12$ layers each with $12$ attention heads, and $768$ hidden units.} We refer to this model as IndT5\textsubscript{100k}. Afterwards, we further pre-train on only the ten Indigenous languages part of our dataset (i.e., without the Spanish data) for $40$K steps. We refer to this version of the model as IndT5\textsubscript{140k}. For both pre-training steps, we use a learning rate of $0.01$, a batch size of $128$ sequences, and a maximum sequence length of $512$. We use the original implementation of T5 in the TensorFlow framework.~\footnote{\url{https://github.com/google-research/text-to-text-transfer-transformer}}.  We train the models on Google Cloud TPU with $8$ cores (v$3.8$) from TensorFlow Research Cloud (TFRC).\footnote{\url{https://www.tensorflow.org/tfrc}}

\section{Our Machine Translation Models}\label{sec:our_MT_models}

\begin{table*}[t]
\small
\centering
\begin{adjustbox}{width=16cm}
\renewcommand{\arraystretch}{1.8}
{
        \begin{tabular}{lll}
        \toprule

      \textbf { Pair }   &\textbf {~~~~~~~~~~~~~~~~~~~~~~~ Sentence } & \textbf{ ~~~~~~~~~~~~~~~~~~~~~~~~~~~~~~~~~~~~~~~~~~~~~~~~Translation} \\    \toprule
            
\multirow{2}{*}{\textbf{es-aym}   } & Algunos actores usan el teatro comunitario para mejorar.   &      \small{ Yaqhip akturanakax juk'amp yatsuñatakiw ayllunkir tiyatrur mantapxi. }\\ 
   & Los artistas de IRT ayudan a los niños en las escuelas.  &      \small{ IRT artistanakax jisk'a yatiqañ utankir wawanakaruw yanapapxi. }\\ 
\hline

 \multirow{2}{*}{\textbf{es-cni}   }   &Pensé que habías ido al campamento.     &      \small{ Nokenkeshireashitaka pijaiti imabeyetinta.}\\ 
   & Viajar es un beneficio que obtenemos.  &      \small{Akenayeeterika aparo ayeeti aneakeri.}\\ 

\hline

\multirow{2}{*}{\textbf{es-bzd}   } & Fui a un seminario que se hizo vía satélite.      &      \small{Ye' dë'rö seminario ã wéx yö' satélite kĩ.}\\ 
   & El grupo está interesado en temas ambientales.   &      \small{E' wakpa kĩ ujtè kiànã e' dör káx ajkóqnũk.}\\ 
\hline

  \multirow{2}{*}{\textbf{es-gn}   } & Veía a su hermana todos los días.     &      \small{Ko'êko'êre ohecha heindýpe.}\\ 
   &  Ramona nunca ha estado en Concord.        &      \small{Ramona noîriva Concord-pe.}\\ 
\hline

\hline
    
 \multirow{2}{*}{\textbf{es-nah}   }   &  Santo trabajó para Disney y operó las tazas de té.      &      \small{ zanto quitequitilih Disney huan quinpexontih in cafen caxitl   }\\ 
  & La hermana de la abuela no era blanca.  &      \small{ihueltiuh in cihtli ixchipahuac catca}\\ 

\hline

\multirow{2}{*}{\textbf{es-quy}   }& De vez en cuando me gusta comer ensalada.    &      \small{Yananpiqa ensaladatam mikuytam munani  }\\ 
   & Ellos vivían en Broad Street.  &      \small{ Broad Streetpi paykuna yacharqaku.  }\\ 
\hline

  \multirow{2}{*}{\textbf{es-tar}   }   & Es un hombre griego. &      \small{Bilé rejói Griego ju }\\ 
  &  Nuestro padre dijo que no los llamaran animales. &      \small{Kini onó aniyé mapu ke chuwé namúti anéba ajaré jákami.}\\ 

\hline

\multirow{2}{*}{\textbf{es-shp}   }& El Museo se ve afectado por las inversiones.   &      \small{Ja Museora en oinai inversionesbaon afectana.  }\\ 
   & Loren Field es el científico principal de la escuela  &      \small{ Nato Loren Field riki científico rekena axeti xobonko  }\\ 
\hline

\multirow{2}{*}{\textbf{es-hch}   }& Era una selva tropical. &      \small{ pe h+k+t+kai metsi+ra+ ye tsie nieka ti+x+kat+.  }\\ 
   & Son más económicos porque son realmente buenos en gas.  &      \small{p+ h+k+ nip+ka raye at+ka aix+ m+ anenek+ ik+ gas.}\\ 
\hline      
    \toprule

\end{tabular}}
\end{adjustbox}
    \caption{Example sentences of the various language pairs and corresponding translations (from Dev set).}
    
    \label{tab:exp}
\end{table*}

\subsection{Parallel Data }
As part of the AmericasNLP 2021 Shared Task on Open Machine Translation, the training (Train) and development (Dev) datasets for ten target Indigeneous languages along with the source language Spanish were released. All the datasets are  manually translated. Table~\ref{tab:ST_data} shows the number of sentences of different language pairs in shared task data.  
Table~\ref{tab:exp} provides example sentences extracted from the Dev dataset with their corresponding translations.

\subsection{Approach}\label{subsec:Approach}
For all languages pairs except \textit{quy} and \textit{gn}, we fine-tune each of the two versions of our language model, i.e., both  IndT5\textsubscript{100k} and IndT5\textsubscript{140k}, under two conditions: 
\textbf{(A)} we train on Train using $100$\% of Dev data for validation, for 150 epochs; \textbf{(B)} we fine-tune the best epoch from setting A for 50 epochs, adding $80$\% of Dev data to Train (using the remaining $20$\% Dev for validation).

\subsection{Evaluation}

We report the results of both IndT5\textsubscript{100k} and IndT5\textsubscript{140k} models using two metrics: BLEU score~\cite{papineni2002bleu} and ChrF++~\cite{popovic2017chrf++}. Tables \ref{tab:res1} and \ref{tab:res2} show the results of both models  on Test sets for each of the language pairs using settings A and B described in Section~\ref{subsec:Approach}, respectively.

\subsection{Discussion}

The results presented in Table~\ref{tab:res1} and Table~\ref{tab:res2}  show that all our models, with both settings A and B, outperform the respective baselines across all languages. An exception is the languages \textit{aym} and \textit{shp}. As expected, fine-tuning the IndT5\textsubscript{100k} and IndT5\textsubscript{140k} models using the training data and $80$\% of the Dev data (i.e., setting B) improves the results with a mean of $+0.003\%$ and  $+0.04\%$ in ChrF++ on the Test data, respectively. Interestingly, further pre-training IndT5 on only the  ten Indigenous languages  (i.e. target languages) produces better results with an average improvement of  $+0.003\%$ and $+0.004\%$  in settings A  and B, respectively. Overall, the impact of limited data is clear.

 \begin{table}[]
\centering
\resizebox{1\columnwidth}{!}{%
\begin{tabular}{lcccccc}
\toprule
\multicolumn{1}{c}{\multirow{2}{*}{\textbf{Pair}}} & \multicolumn{2}{c}{\textbf{\textbf{Baseline}  }} &\multicolumn{2}{c}{\textbf{\textbf{Setting A}  }} & \multicolumn{2}{c}{\textbf{ \textbf{Setting B} }}\\  \cline{2-7}
 &\textbf{ Bleu} & \textbf{ChrF++ }  &   \textbf{ Bleu} & \textbf{ChrF++ } &  \textbf{Bleu} & \textbf{ChrF++ }    \\ \toprule

\textbf{aym} & 0.3  &\textbf{ 0.188 }& 1.01                  & 0.178                 & 0.76                  & 0.186                 \\
\textbf{cni} & 0.03 & 0.104 & 0.09                  & 0.176                 & 0.09                  & \textbf{0.178 }                \\
\textbf{bzd} & 0.54 & 0.077 & 0.86                  & 0.11                  & 0.89                  & \textbf{0.111  }               \\
\textbf{oto} & 0.01 & 0.059 & 0.03                  & 0.081                 & 0.04                  &\textbf{ 0.083   }              \\
\textbf{nah} & 0.33 & 0.182 & \multicolumn{1}{c}{-} & \multicolumn{1}{c}{-} & 0.16                  & \textbf{0.196   }              \\
\textbf{tar} & 0.01 & 0.046 & 0.06                  & 0.102                 & -                     & -                     \\
\textbf{hch} & 3.18 & 0.126 & 4.95                  & \textbf{0.186}                 & 5.09                  & \textbf{0.186 }          \\      
\toprule 
\end{tabular}}
\caption{Evaluation results of IndT5\textsubscript{100k} in BLEU and ChrF++ on the Test  sets for the different language pairs.}  \label{tab:res1}

\end{table}

 \begin{table}[]
\centering
\resizebox{1\columnwidth}{!}{%
\begin{tabular}{lcccccc}
\toprule
\multicolumn{1}{c}{\multirow{2}{*}{\textbf{Pair}}} & \multicolumn{2}{c}{\textbf{\textbf{Baseline}  }} &\multicolumn{2}{c}{\textbf{\textbf{Setting A}  }} & \multicolumn{2}{c}{\textbf{ \textbf{Setting B} }}\\  \cline{2-7}
 &\textbf{ Bleu} & \textbf{ChrF++ }  &   \textbf{ Bleu} & \textbf{ChrF++ } &  \textbf{Bleu} & \textbf{ChrF++ }    \\ \toprule

\textbf{aym} & 0.3  & 0.188 & 0.820 & 0.182 & 0.990 &\textbf{ 0.190} \\
\textbf{cni} & 0.03 & 0.104 & 0.070 & 0.178 & 0.080 & \textbf{0.183} \\
\textbf{bzd} & 0.54 & 0.077 & 0.990 & 0.112 & 0.940 & \textbf{0.113} \\
\textbf{oto} & 0.01 & 0.059 & 0.030 & 0.082 & 0.040 & \textbf{0.08}4 \\
\textbf{nah} & 0.33 & 0.182 & 0.150 & 0.188 & 0.160 &\textbf{ 0.196} \\
\textbf{tar} & 0.01 & 0.046 & 0.080 & 0.102 & 0.050 & \textbf{0.10}5 \\
\textbf{shp} & 0.34 &\textbf{ 0.139} & 0.160 & 0.124 & 0.230 & 0.124 \\
\textbf{hch} & 3.18 & 0.126 & 5.100 & 0.194 & 5.520 & \textbf{0.195}         \\    
\toprule 
\end{tabular}}
\caption{Evaluation results of IndT5\textsubscript{140k} in BLEU and ChrF++ on the Test sets for the different language pairs.}  \label{tab:res2}

\end{table}

\section{Conclusion}\label{sec:conc}
In this work, we introduced a new Transformer language model (\textbf{\texttt{IndT5}}) and a dataset (\textbf{\texttt{IndCorpus}}) for ten Indigenous languages and Spanish. We applied \textbf{\texttt{IndT5}} to the MT task on eight languages pairs as part of our submission to the AmericasNLP 2021 Shared Task.  While \textbf{\texttt{IndT5}} helps improve translation, the task remains hard due to absence of parallel as well as monolingual data. In the future, we plan to integrate statistical MT methods to augment our data as well as investigate best hyperparameters for our neural models.

\vspace{2mm}
\section*{Acknowledgements}\label{sec:acknow}

\vspace{2mm}
We gratefully acknowledge support from the Natural Sciences and Engineering Research Council of Canada, the Social Sciences and Humanities Research Council of Canada, Canadian Foundation for Innovation, Compute Canada, and UBC ARC-Sockeye. We also thank the Google TFRC program for providing us with free TPU access.\\

\bibliography{anthology,custom}
\bibliographystyle{acl_natbib}

\appendix



\end{document}